\documentclass[letterpaper]{article} 
\usepackage{aaai2026}  
\usepackage{times}  
\usepackage{helvet}  
\usepackage{courier}  
\usepackage[hyphens]{url}  
\usepackage{graphicx} 
\usepackage{natbib}  
\usepackage{caption} 
\usepackage{algorithm}
\usepackage{algorithmic}
\usepackage{multirow,multicol}
\usepackage{booktabs}
\usepackage{amsmath}
\usepackage{cleveref}

\usepackage{newfloat}
\usepackage{listings}
\DeclareCaptionStyle{ruled}{labelfont=normalfont,labelsep=colon,strut=off} 
\floatstyle{ruled}
\newfloat{listing}{tb}{lst}{}
\floatname{listing}{Listing}
\title{DASH: Dialogue-Aware Similarity and Handshake Recognition for Topic Segmentation in Public-Channel Conversations}
\author{
Sun Sijin\textsuperscript{\rm 1,3},
Liangbin Zhao\textsuperscript{\rm 1,*},
Ming Deng\textsuperscript{\rm 2},
Xiuju Fu\textsuperscript{\rm 1}
}
\affiliations{
\textsuperscript{\rm 1}Institute of High Performance Computing, Agency for Science Technology and Research (A*STAR IHPC) \\
\textsuperscript{\rm 2}Shanghai University \\
\textsuperscript{\rm 3}National University of Singapore
}

\usepackage{bibentry}

\begin{document}

\maketitle

\begin{abstract}

Dialogue Topic Segmentation (DTS) is crucial for understanding task-oriented public-channel communications, such as maritime VHF dialogues, which feature informal speech and implicit transitions. To address the limitations of traditional methods, we propose DASH-DTS, a novel LLM-based framework. Its core contributions are: (1) topic shift detection via dialogue handshake recognition; (2) contextual enhancement through similarity-guided example selection; and (3) the generation of selective positive and negative samples to improve model discrimination and robustness. Additionally, we release VHF-Dial, the first public dataset of real-world maritime VHF communications, to advance research in this domain. DASH-DTS provides interpretable reasoning and confidence scores for each segment. Experimental results demonstrate that our framework achieves several sota segmentation trusted accuracy on both VHF-Dial and standard benchmarks, establishing a strong foundation for stable monitoring and decision support in operational dialogues.

\end{abstract}

\begin{links}
    \link{Code}{https://github.com/StanleySun233/dash-dts}
    \link{Datasets}{https://github.com/StanleySun233/dash-dts/dataset/vhf.json}
    \link{Demos}{https://github.com/StanleySun233/dash-dts/blob/main/demo/demo.gif}
\end{links}

\section{Introduction}
Understanding topic structure in dialogue is essential for a wide range of downstream tasks, including summarization \cite{han2024let}, event detection \cite{davies2009discussing}, dialogue planning, and regulatory decision support. At the core of such understanding lies the task of Dialogue Topic Segmentation (DTS)—the process of identifying boundaries where the conversation transitions from one topic to another \cite{zhang2019topic}. While considerable progress has been made in open-domain and customer-service dialogue segmentation \cite{artemiev2024leveraging}, public-channel conversations—such as those in maritime Very High Frequency (VHF) radio, air traffic control, or emergency dispatch networks—remain underexplored despite their societal and operational importance.

Public-channel dialogues differ fundamentally from traditional open-domain \cite{feng2021survey} or daily conversations. They are characterized by extremely short, fragmented utterances, dynamically shifting speaker roles, and high-stakes operational intent. In such settings, topic transitions are often implicit, driven by speaker intent and interactional coordination rather than explicit lexical cues or discourse markers. These structural and pragmatic characteristics challenge the assumptions of existing DTS models, which typically rely on surface-level continuity, utterance embedding similarity, or turn-level encoders—while largely overlooking speaker-driven signals that indicate topical shifts.

These challenges are not merely academic. In industrial maritime regulation, for instance, analysts are often tasked with reviewing hours of VHF voice communications to identify key events such as near-miss collisions or miscommunications. These public-channel interactions, while rich in safety-critical information, are rarely structured or archived due to their volume and fragmented nature. In practice, identifying task shifts, escalation cues, or behavioral patterns still heavily relies on manual transcription and expert judgment. This bottleneck not only limits the coverage of maritime oversight systems but also results in missed opportunities to learn from near-miss cases—which are far more frequent and informative than actual accidents. A robust DTS system tailored for public-channel dialogues \cite{gao2023unsupervised} would drastically reduce this cost by offering structure-aware segmentation that supports downstream analytics, compliance auditing, and real-time alerting.

Existing DTS approaches primarily rely on surface-level lexical transitions or embedding similarity between utterances, often using cosine similarity to retrieve relevant examples. While such methods work reasonably well in structured or open-domain settings, they struggle in task-oriented, public-channel dialogues—where utterances are terse, speaker roles shift dynamically, and topic shifts are often implicit. In particular, these conversations frequently contain short, functional “dialogue handshakes”—such as “Star Alpha calling port control”—which act as subtle signals of upcoming topical change. However, such interactional patterns are rarely modeled in existing systems. Meanwhile, in-context learning (ICL) with large language models \cite{rubin2021learning} offers a promising alternative for few-shot topic segmentation, but selecting semantically appropriate exemplars remains challenging in domain-specific, sparse-data environments like VHF communication.

To address these challenges, we propose DASH-DTS (Dialogue-Aware Similarity and Handshake recognition for Dialogue Topic Segmentation), a structure-aware framework for segmenting topics in public-channel conversations. DASH-DTS incorporates three core components: (1) a handshake recognition module that identifies short interactional cues marking the onset of new topical segments; (2) a dialogue similarity-guided in-context learning strategy, which retrieves semantically relevant exemplars to enhance segmentation in sparse-data conditions; and (3) a context-aware labeling mechanism that utilizes surrounding discourse to produce more coherent and accurate topic annotations. Besides, to support trustworthy deployment in the follow-up applications, our framework additionally generates segment-level justifications and confidence scores, allowing users to assess the reliability of predicted topic boundaries. 

In addition, to support research and benchmarking in this domain, we construct and release the first publicly available DTS dataset for maritime VHF communications. This dataset captures the unique characteristics of public-channel dialogues—such as brief utterances, implicit transitions, and dynamic speaker roles—and serves as a valuable resource for evaluating topic segmentation methods in real-world, safety-critical communication environments.

\noindent The main contributions of this work are as follows:
\begin{itemize}
    \item Propose a handshake recognition mechanism that captures speaker interaction cues to identify topic boundaries in public-channel dialogue, addressing structural challenges in dialogue topic segmentation.
    
    \item Introduce a similarity-guided in-context learning strategy that selects semantically relevant exemplars to enhance segmentation performance for DTS in sparse and domain-specific settings.

    \item Introduce an interpretable and trustworthy output mechanism that generates segment-level justifications and confidence scores, enabling downstream applications to assess the reliability of topic boundaries.

    \item Construct and release the first publicly available dataset for dialogue topic segmentation in the VHF public-channel communication domain, termed \textbf{VHF-Dial}, providing a benchmark for real-world application.

\end{itemize}

\section{Related Work}

\paragraph{Early Methods} Traditional approaches to Dialogue Topic Segmentation (DTS) primarily relied on lexical cohesion and surface-level continuity. Techniques like TextTiling \cite{hearst1997texttiling} segmented text by identifying lexical valleys in cosine similarity between adjacent blocks. While effective for monologic texts, these methods struggle with dialogues due to their fragmented utterances, speaker shifts, and informal language \cite{misra2009text}. For task-oriented dialogues, where topics shift implicitly without explicit lexical cues, such approaches exhibit significant limitations in robustness \cite{song2016dialogue}.

\paragraph{Sequence Modeling Methods} To capture contextual dependencies, later works adopted sequence modeling architectures. SimCSE \cite{gao2021simcse} models leveraged utterance representations to predict topic boundaries sequentially, improving coherence modeling. However, these methods often require extensive labeled data and fail to generalize across diverse conversational domains. Their reliance on local context also overlooks global dialogue structure.

\paragraph{Pre-trained Models} The advent of PLMs like BERT revolutionized DTS by enabling deeper semantic understanding. Baseline evaluate of DTS \cite{7} fine-tuned BERT for utterance-pair coherence scoring, using topical relevance as a segmentation signal. Similarly, unsupervised frameworks like Topic-aware Utterance Representation \cite{artemiev2024leveraging} leveraged pseudo-segmentation tasks on unlabeled dialogues. While PLMs improved accuracy, they remain constrained by domain transferability issues and high computational costs \cite{10}, particularly in low-resource, noisy environments (e.g., emergency dispatch systems).

\paragraph{Large Language Models} Recent efforts integrate LLMs for few-shot segmentation. DEF-DTS \cite{10} employs multi-step deductive reasoning via structured prompting. S3-DST \cite{das2024s3dst} uses self-supervised learning to mitigate data scarcity. Despite promising results, LLM-based methods still face challenges on limited interpretability of topic transitions, and neglect of structural dynamics \cite{12}.

Unlike prior work, DASH-DTS uniquely integrates dialogue handshake recognition to detect structural cues and semantic pivots for topic shift detection. We avoid LLMs' prompt engineering limitations by leveraging similarity-based in-context learning with dynamically retrieved examples. Our framework is specifically designed for noisy, high-stakes dialogue, where implicit transitions and dynamic speaker roles invalidate assumptions of existing methods \cite{10}. The release of a curated maritime DTS dataset further addresses chronic data scarcity in this domain.

\begin{figure*}
    \centering
    \includegraphics[width=1\linewidth]{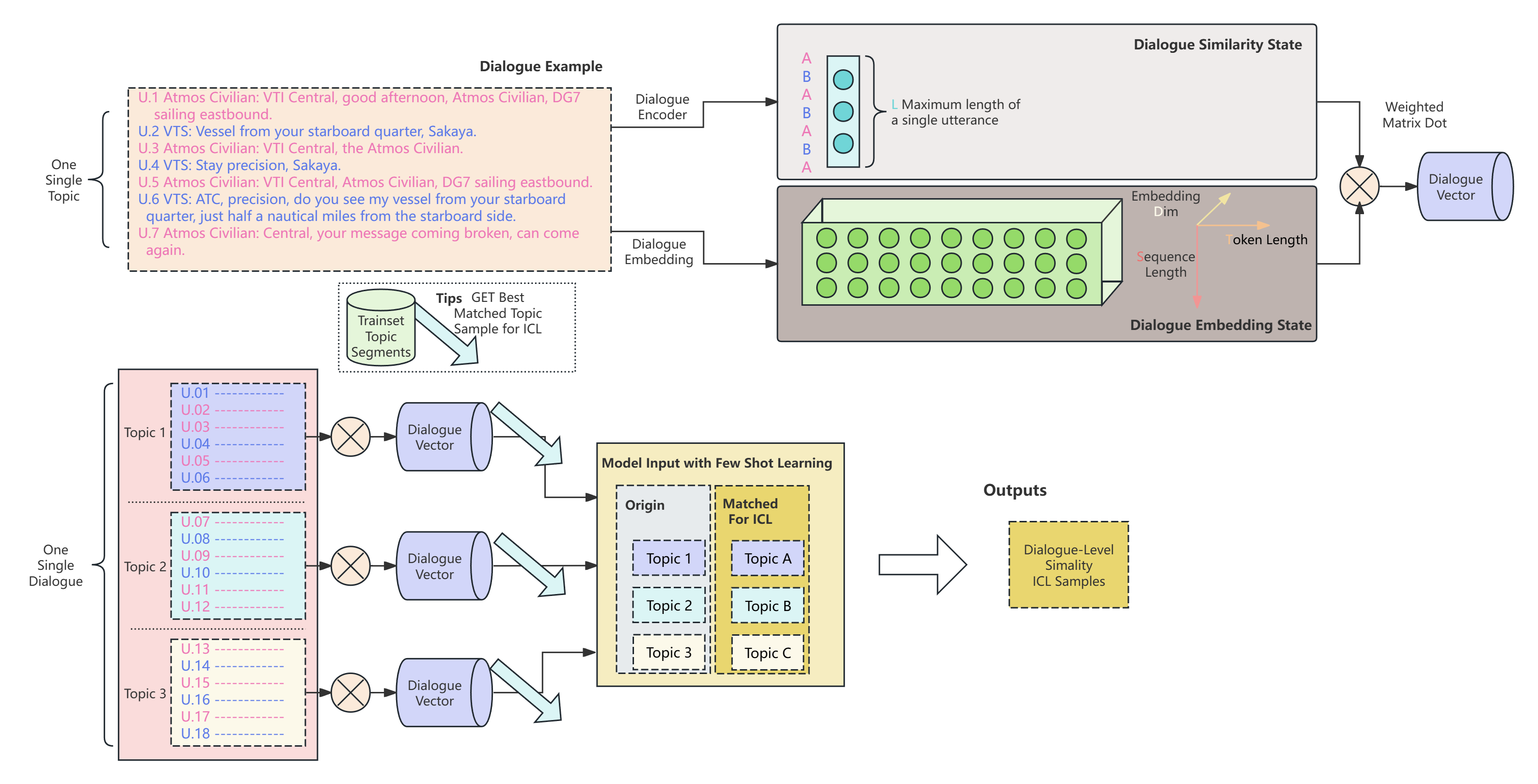}
    \caption{Dynamically select the most semantically relevant exemplars for each input conversation, enhancing the accuracy of topic segmentation in contextual learning for large models.}
    \label{fig:dlsicl}
\end{figure*}

\section{Methodology}
\subsection{Dialogue-level Similarity}
To effectively leverage in-context learning (ICL) in the domain of public-channel dialogue topic segmentation, it is crucial to select semantically relevant examples that can guide the model towards more accurate segmentation. Traditional ICL approaches often rely on cosine similarity or surface-level lexical matching, which may not be sufficient in scenarios with sparse and highly specialized data like VHF maritime communications. Therefore, we introduce a novel dialogue similarity-guided in-context learning strategy to enhance the performance called Dialogue-Level Similarity ICL Samples in \Cref{fig:dlsicl}.

\paragraph{Semantic Similarity Calculation} The first step in our approach is to compute the semantic similarity \cite{lavi2021we} between the query dialogue and the available exemplars. We use a pre-trained language model, such as BERT or RoBERTa, fine-tuned on a large dialogue dataset, to generate embeddings for each utterance. These embeddings capture the contextual and semantic information of the utterances \cite{abro2022natural}, making them more suitable for comparing the underlying meanings rather than just the surface text.

Given a query dialogue $D_q$ and a set of exemplars $E = \{E_1, E_2, \ldots, E_n\}$, we compute the embedding for each utterance in $D_q$ and $E_i$. Let $U_{q,j}$ be the $j$-th utterance in $D_q$ and $U_{i,k}$ be the $k$-th utterance in $E_i$. The embeddings are denoted as $Emb(U_{q,j})$ and $Emb(U_{i,k})$.

We then calculate the pairwise cosine similarity between the utterances in the query dialogue and the exemplars:

\begin{equation}
    \text{sim}(U_{q,j}, U_{i,k}) = \frac{Emb(U_{q,j}) \cdot Emb(U_{i,k})}{\|Emb(U_{q,j})\| \|Emb(U_{i,k})\|}
\end{equation}

Next, we aggregate the similarities at the dialogue level. One way to do this is by averaging the cosine similarities of all utterance pairs:

\begin{equation}
    \text{sim}(D_q, E_i) = \frac{1}{|D_q| \times |E_i|} \sum_{j=1}^{|D_q|} \sum_{k=1}^{|E_i|} \text{sim}(U_{q,j}, U_{i,k})
\end{equation}

Alternatively, we can use a weighted aggregation method, where the weights are determined by the importance of each utterance in the dialogue. For example, we can assign higher weights to utterances that are more likely to indicate a topic change, such as those containing dialogue handshakes.

\paragraph{Exemplar Selection} Once we have computed the semantic similarity between the query dialogue and the exemplars, we select the most relevant exemplars to include in the context. We rank the exemplars based on their similarity scores and choose the top $m$ exemplars, where $m$ is a hyperparameter that can be tuned based on the specific application and available computational resources.

\begin{equation}
    E_{\text{selected}} = \arg\max_{E_i \in E} \text{sim}(D_q, E_i)
\end{equation}

These selected exemplars are then concatenated with the query dialogue to form the input for the in-context learning process. The model uses these exemplars to learn the patterns and structures that are indicative of topic transitions in the given dialogue.

\paragraph{In-Context Learning with Selected Exemplars}
With the selected exemplars, we perform in-context learning using a large language model. The concatenated input consists of the selected exemplars followed by the query dialogue. The model is prompted to predict the topic boundaries in the query dialogue, guided by the examples provided in the context.

\begin{equation}
    \text{Input} = [E_{\text{selected}}; D_q]
\end{equation}

The model outputs a sequence of labels indicating the topic boundaries in $D_q$. By providing semantically relevant exemplars, the model can better generalize to the query dialogue, even in the presence of sparse and specialized data.

\subsection{Handshake Statement Tag}

In public-channel dialogues, particularly maritime VHF communications, "handshake" statements function as critical interactional markers that demarcate topical boundaries. These statements typically comprise concise, functionally-oriented utterances signaling conversational focus shifts or topic transitions \cite{konigari2021topic}. Representative examples include phrases such as "Star Alpha calling port control" or "Delta Echo, this is Bravo Hotel," which serve as subtle cues facilitating topic segmentation. Accurate identification of these statements is essential for robust topic segmentation, especially in high-stakes communication environments where implicit topic transitions and dynamic speaker roles introduce additional complexity.

\begin{figure}[!htbp]
    \centering
    \includegraphics[width=1\linewidth]{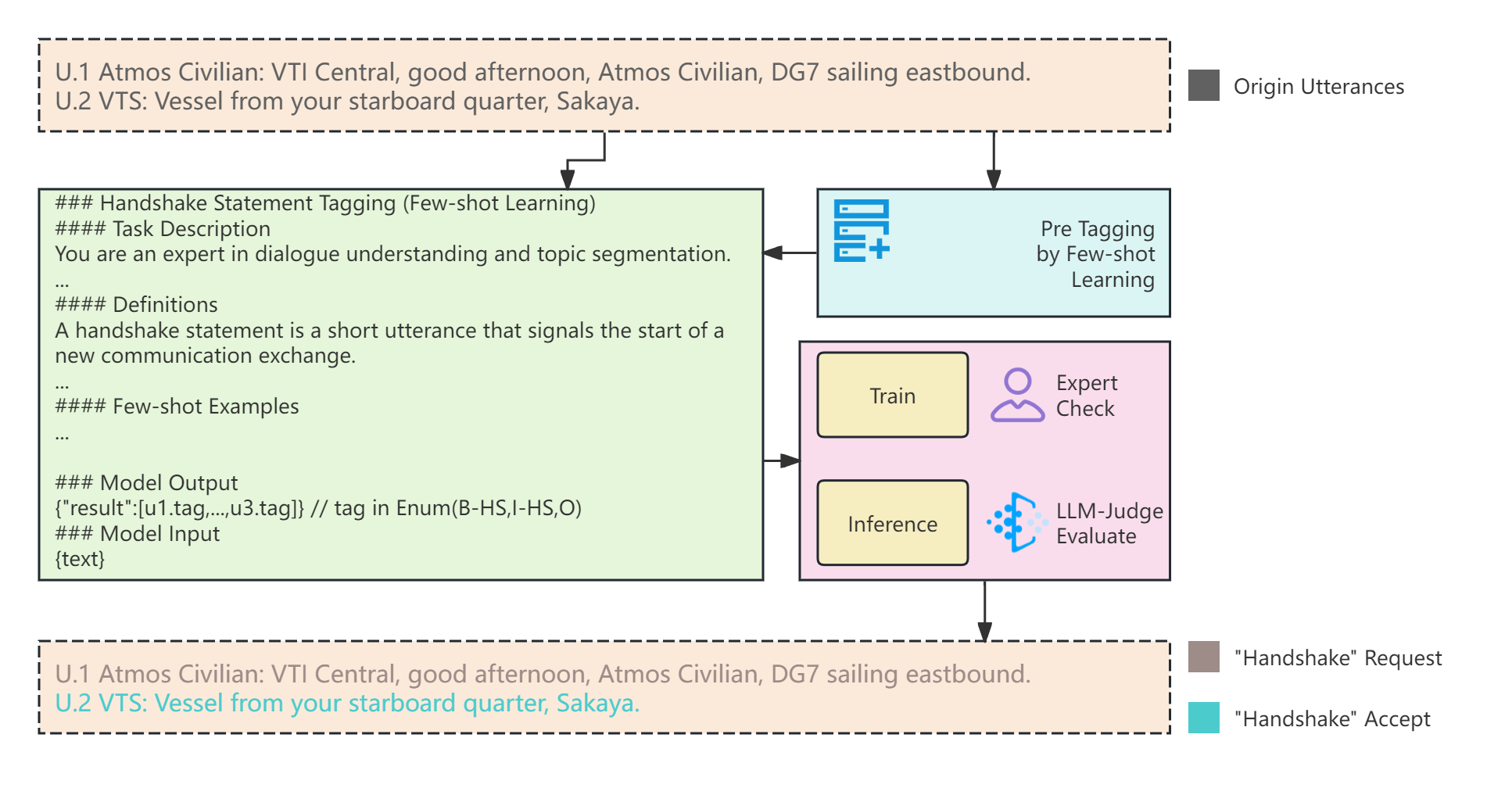}
    \caption{Workflow of the Handshake Statement Tagging Component Based on Few-shot LLM Learning.}
    \label{fig:handshake}
\end{figure}

Given a dialogue sequence $D = \{U_1, U_2, \ldots, U_n\}$ where each $U_i$ represents an utterance decomposed into tokens, the handshake identification task is formulated as a token-level sequence labeling problem. Each token is assigned one of three labels: (1)HS-BEG: Beginning of a handshake statement, (2) HS-END: End of a handshake statement, (3) O: Inside or indeterminate portions of dialogues.

To systematically identify handshake statements, the Handshake (HS) Agent is introduced as a reusable component within the DASH-DTS framework, as illustrated in \Cref{fig:handshake}. The HS agent leverages the capabilities of large language models to recognize these interactional cues \cite{brysbaert2022marking}. Rather than relying solely on deterministic pattern matching, the HS agent is designed to produce structured, interpretable outputs that facilitate both automated decision-making and human verification.

The HS agent processes dialogue utterances and generates structured predictions for each token. Each prediction is formulated as a principled triplet in \Cref{eq:hs-structure}.

\begin{equation}
    P_i=(l_i,s_i,r_i) 
    \label{eq:hs-structure}
\end{equation}
where $l_i \in \{\text{HS-BEG, HS-END, O}\}$ denotes the predicted label, $s_i \in [0,1]$ represents the trustworthiness score indicating prediction confidence, and $r_i$ encapsulates the reasoning justification underlying the classification decision.

This structured restriction ensures that predictions are not only accurate but also transparent and auditable. The reasoning component $r_i$ documents the linguistic and contextual evidence supporting each classification, thereby enabling domain experts to assess prediction reliability and identify potential failure modes.

For each dialogue, the agent is prompted with contextual specifications and exemplars of typical handshake statement patterns. The LLM component generates label predictions for token sequences, accompanied by explicit trustworthiness assessments and reasoning justifications. Post-processing constraints are applied to ensure label coherence: each HS-BEG label must be paired with a corresponding HS-END label, maintaining proper demarcation of handshake statement boundaries. This step enforces structural consistency and eliminates malformed predictions. The final output comprises:

\begin{itemize}
    \item Token-level label sequence: Predictions of handshake statement boundaries (onset and termination points)
    \item Trustworthiness scores: Confidence metrics reflecting prediction reliability for each token
    \item Reasoning chains: Interpretable justifications documenting the evidence supporting each classification decision
\end{itemize}

Proposed handshake statement predictions, enriched with trustworthiness scores and reasoning chains, are integrated into the DASH-DTS framework to identify structural cues signaling the onset of new topical segments. This principled integration enhances both the robustness and interpretability of the overall topic segmentation process, particularly in scenarios characterized by implicit transitions and dynamic speaker configurations.

\subsection{Content-Aware Topic Generation}

For enhancing the in-context learning (ICL) process and improve the accuracy of topic segmentation, we propose a content-aware topic generation mechanism. This mechanism leverages the full context of the dialogue to generate positive and negative samples, which are then used to guide the model in identifying potential topic boundaries. The key idea is to create synthetic examples that help the model understand the characteristics of both topic transitions and non-transitions.

Given a dialogue segment with contextual window $W = \{U_{i-m}, \ldots, U_i | U_{i+1}, \ldots, U_{i+n}\}$ where $U_i$ and $U_{i+1}$ form a potential segmentation point, we formulate the sample generation task as a structured prediction problem. Let $\mathcal{S} = (D_p, D_n, \mathcal{E})$ denote the generated sample triplet, where $D_p = \{s_1^p, s_2^p, \ldots, s_7^p\}$ is a positive dialogue sample with ground truth label $y_p = 1$, $D_n = \{s_1^n, s_2^n, \ldots, s_7^n\}$ is a negative dialogue sample with ground truth label $y_n = 0$, and $\mathcal{E} = \{\xi_p, \xi_n, \mathcal{C}\}$ represents explainability artifacts including reasoning chains and confidence scores. The core challenge is to automatically generate balanced, diverse samples while maintaining semantic coherence and providing verifiable reasoning traces that enable human inspection.

\begin{table*}[!htbp]
\centering
\caption{Experimental Results on Our Conducted Datasets of public VHF channel dialog}
\begin{tabular}{llcccccc}
\toprule
\multicolumn{1}{c}{\multirow{2}{*}{Model}} & \multicolumn{1}{c}{\multirow{2}{*}{Reference}} & \multicolumn{2}{c}{DialSeg711} & \multicolumn{2}{c}{Doc2Dial} & \multicolumn{2}{c}{VHF-Dial} \\
\cmidrule(lr){3-4} \cmidrule(lr){5-6} \cmidrule(lr){7-8}
\multicolumn{1}{c}{}                       & \multicolumn{1}{c}{}                        & $P_k$           & $W_d$           & $P_k$           & $W_d$           & $P_k$           & $W_d$         \\ 
\midrule
Text Tiling                               & \cite{hearst1997texttiling}                  & 40.4        & 44.6        & 52.0         & 57.4        & 54.3       & 61.7       \\
LLM & gemini-2.5-flash & 36.5 & 67.9 & 46.4 & 53.5 & 38.6 & 78.6 \\
DyDTS                                      & \cite{Lv2025DynamicTS}             & 24.7        & \textbf{27.6}        & 39.9         & 44.0        & 38.2       & 39.7       \\
UPS                                        & \cite{ups}               & --            & --            & 35.1          & 36.5         & 27.4       & 34.5       \\
SumSeg                                     & \cite{artemiev2024leveraging}               & 47.7         & 48.3         & --             & --            & 32.7       & 35.1       \\
CSM                                       & \cite{gao2023unsupervised}                   & 26.8        & 28.2         & 45.2             & 47.3           & 27.7       & 31.6       \\
BERT                                       & \cite{bert}               & 39.3        & 41.2        & 53.7         & 55.3        & 44.9       & 49.1       \\ 
\midrule
ours                                       &                                             & \textbf{20.7}         & 34.3         & \textbf{33.9}         & 36.6        & \textbf{21.9}       & \textbf{33.9}       \\ 
\bottomrule
\end{tabular}
\label{tab:comp}
\end{table*}

\paragraph{Contextual Input} Given the contextual window $W$, the $L$ first performs deep semantic analysis of the dialogue through a structured analysis. This analysis stage instructs the model to extract thematic elements and discourse topics from each utterance, identify lexical and pragmatic markers such as discourse particles and topic shift indicators, characterize speaker roles and dialogue coherence patterns, and detect domain-specific terminologies and contextual dependencies. The output $\mathcal{A}$ is a structured analysis document that serves as grounding for subsequent stages, ensuring that sample generation is semantically faithful to the original dialogue and reflects the actual discourse structure rather than superficial patterns.

\begin{equation}
\mathcal{A} = \text{L}_{\theta}(\text{analyze}(W) \mid \tau_{\text{analysis}})
\end{equation}

Then, building on the analysis $\mathcal{A}$, we perform dual-mode generation structure ($\mathcal{GC}$) to create positive and negative samples through contrastive synthesis. 

\begin{equation}
(D_p, D_n) = \text{LLM}_{\theta}(\mathcal{GC}(W, \mathcal{A}) \mid \tau_{\text{synthesis}})
\end{equation}

\paragraph{Posi/Nega-tive Generation} For positive samples, the synthesis prompt instructs the LLM to generate utterances that maintain thematic continuity in the previous segment (positions 1-3), create a pivot utterance at position 4 that exhibits explicit topic shift markers including transitional phrases such as ``By the way'' or ``Speaking of'', sudden perspective changes, or domain switches, and extend with utterances in the next segment (positions 5-7) that cohere around the new topic. The generated dialogue should present a clear and unambiguous boundary that enables the model to learn definitive topic transition signals.

For negative samples, the synthesis prompt requires the LLM to maintain strong thematic and pragmatic continuity across all seven positions, ensure that the pivot utterance advances, elaborates on, or clarifies the previous topic rather than shifting away, and employ within-topic discourse patterns such as agreement, clarification, and detail addition rather than boundary markers. These negative examples are particularly important as they teach the model to distinguish between genuine topic boundaries and discourse continuations that might superficially resemble transitions.

Crucially, both synthesis operations preserve the stylistic, register, and speaker role patterns observed in $\mathcal{A}$, ensuring that $D_p$ and $D_n$ remain authentic exemplars of real dialogue phenomena rather than synthetic artifacts that might introduce spurious patterns.

The complete output is structured to provide comprehensive traceability:
\begin{equation}
\mathcal{S} = \begin{cases}
\text{positive}: \{D_p, y_p=1, \mathcal{C}_p, \xi_p\} \\
\text{negative}: \{D_n, y_n=0, \mathcal{C}_n, \xi_n\}
\end{cases}
\end{equation}

\paragraph{Trustworthy CoT} This multi-layered output format offers several critical advantages for trustworthy AI systems. The reasoning traces $\xi_p, \xi_n$ provide human-readable justifications that enable auditors to verify that samples reflect intended classification logic rather than spurious correlations learned from training data. The confidence scores $\mathcal{C}_p, \mathcal{C}_n$ enable downstream systems to weight samples appropriately during training, downweighting low-confidence exemplars and preventing overfitting to unreliable signals. When the topic segmentation model makes errors during deployment, practitioners can inspect the explanations to diagnose whether errors stem from inadequate sample generation or model learning failures, supporting systematic debugging and improvement. By explicitly recording the reasoning chains and confidence scores, our method creates an auditable paper trail suitable for regulatory compliance and scientific reproducibility, addressing key concerns in trustworthy machine learning.

\subsection{Segment Reliability and Explanation}
To enhance the trustworthiness and usability of topic segmentation in high-stakes scenarios, DASH-DTS outputs a confidence score for each predicted segment, reflecting the model's certainty in its boundary decisions. Additionally, a brief natural language explanation is generated to justify each segmentation point, allowing users to better interpret and assess the system's output. This design supports human-in-the-loop workflows and enables more informed downstream decision-making.

\paragraph{Prompt example.}
To support interpretability and trust calibration, we prompt the LLM to provide an explanatory rationale and a confidence score for each predicted topic segment:

\begin{quote}
\small
\texttt{For each predicted topic segment, provide:}\\
(1) A brief explanation of the topical focus or transition, and whether it constitutes a complete dialogue task.\\
(2) A confidence score between 0 and 1 indicating the model's certainty.

\texttt{Example:}

Explanation: This segment involves a vessel initiating contact with port control to request entry clearance. The conversation forms a self-contained exchange where the intent, response, and acknowledgment are completed, indicating a coherent dialogue unit with a clear operational focus.

Confidence: 0.91
\end{quote}

\section{Experiment}
\subsection{Comparison Study}

The metrics used for evaluation are $P_k$\cite{hearst1997texttiling} in \Cref{eq:pk} and $W_d$ (Windo$W_d$iff)\cite{Pevzner2002ACA} in \Cref{eq:wd}, which are standard measures for assessing the quality of topic segmentation.

\begin{equation}
P_k=\frac{\sum_{i=1}^{N-k}\delta_P(i,i+k)\oplus\delta_R(i,i+k)}{N-k}
    \label{eq:pk}
\end{equation}
where $k$ is the window size, $\delta_R$ and $\delta_P$ are indicator functions for a boundary in the reference and prediction, respectively, and $\oplus$ denotes the XOR operation that detects a disagreement.

\begin{multline}
W_d(ref, hyp) = \frac{1}{N-k}\\
\quad \sum_{i=1}^{N-k} (|b(ref_i, ref_{i+k}) - b(hyp_i, hyp_{i+k})| > 0)
\label{eq:wd}
\end{multline}
where $b(i, j)$ represents the number of boundaries between positions $i$ and $j$ in the text, $N$ represents the number of sentences in the text, and $k$ is the window size parameter. This metric compares the number of reference segmentation boundaries with the number of hypothesized boundaries within each sliding window, penalizing the algorithm when these counts differ.

The experimental results, as presented in \Cref{tab:comp}, provide a comprehensive comparison of our proposed DASH-DTS method with several state-of-the-art baselines on three public datasets: \textbf{DialSeg711}, \textbf{Doc2Dial}, and proposed \textbf{VHF-Dial}.

Our proposed VHF-Dial dataset demonstrates distinct challenges and opportunities for dialogue segmentation models. As evidenced in Table \ref{tab:comp}, our method significantly outperforms all baselines on this dataset, achieving a $P_k$ of 21.9 and $W_d$ of 33.9, which represent notable improvements over the best-performing baseline, UPS, accounting for $P_k$ at 27.4. This suggests that our approach is particularly well-suited to the unique characteristics of VHF (Very High Frequency) channel dialogues, which may involve shorter turns, domain-specific jargon, or structured communication protocols.

On the DialSeg711 dataset, our method achieves a $P_k$ score of 20.7, which is 4 percent lower than most baselines of DyDTS. This indicates that while our method captures some topic segments correctly, it may have a higher rate of false positives or over-segmentation. However, our method exhibits room for improvement on the Doc2Dial dataset, which may be attributed to the inherent characteristics of the dataset. This discrepancy suggests that dataset-specific features, such as document structure or dialogue complexity, may influence segmentation effectiveness. Future work could explore adaptive mechanisms to better handle such variations.

\subsection{Ablation Study}
To evaluate the contribution of each component to the performance of DASH-DTS, we conducted ablation experiments on the \textbf{VHF-Dial} in \Cref{tab:ab}. In No.1, we included both the Handshake and Dialogue Similarity components but excluded the Topic Generation component. This configuration resulted in a $P_k$ score of 26.5 and a $W_d$ score of 39.6, indicating that while these two components alone can achieve reasonable performance, the absence of Topic Generation slightly degrades the overall segmentation quality. In No.2, we included only the Dialogue Similarity and Topic Generation components, excluding the Handshake component, which yielded a $P_k$ score of 27.1 and a $W_d$ score of 39.7. This suggests that Dialogue Similarity and Topic Generation are crucial for segmenting the dialogue, but the lack of Handshake recognition leads to a slight decrease in performance.

\begin{table}[!htbp]
\centering
\caption{Ablation Experiment on VHF-Dial}
\resizebox{\columnwidth}{!}{
\begin{tabular}{cccccc}
\toprule
\multicolumn{1}{c}{\multirow{2}{*}{No.}} & \multicolumn{3}{c}{Component}                                             & \multirow{2}{*}{$P_k$} & \multirow{2}{*}{$W_d$} \\
\cmidrule(lr){2-4}
\multicolumn{1}{c}{}                    & Handshake            & Dialogue Similarity & Topic Generation             &                     &                     \\ 
\midrule
1                  & $\surd$ & $\surd$             & & 26.5                 & 39.6                \\
2                  &                      & $\surd$             & $\surd$ & 27.1                 & 39.7                \\
3                  & $\surd$              &                     & $\surd$ & 24.3                 & 34.7                \\ 
\midrule
Ours               &       $\surd$               &        $\surd$             & $\surd$        & \textbf{21.9}                & \textbf{33.9}                \\ 
\bottomrule
\end{tabular}
}
\label{tab:ab}
\end{table}

In No.3, we included the Handshake and Topic Generation components but excluded the Dialogue Similarity component, resulting in a $P_k$ score of 24.3 and a $W_d$ score of 34.7. The removal of Dialogue Similarity led to a higher $W_d$ score, highlighting its importance in maintaining the alignment of segment boundaries. Finally, the full model (Ours) with all three components achieved the best performance, with a $P_k$ score of 21.9 and a $W_d$ score of 33.9. This indicates that the combination of all three components is necessary to achieve optimal topic segmentation, as they collectively contribute to detecting structural cues, enhancing in-context learning, and ensuring semantic coherence.

\subsection{Discussion}

Experimental results provide valuable insights into the performance and contributions of our proposed DASH-DTS method. Here, we discuss three key points based on the findings from the comparison and ablation studies.

1) The ablation study highlights the importance of the handshake recognition mechanism in detecting structural cues and identifying topic boundaries. When the Handshake component is included, the model achieves a lower $W_d$ score, indicating better alignment with the true segment boundaries. This is particularly evident in the DialSeg711 and VHF-Dial, where the absence of Handshake recognition leads to higher $W_d$ scores. The Handshake mechanism effectively captures speaker interaction cues, which are crucial for accurately segmenting topics in public-channel dialogues. This contribution addresses the structural challenges in dialogue topic segmentation, making the model more robust in real-world communication settings.

2) The inclusion of the Dialogue Similarity component significantly enhances the model's performance by selecting semantically relevant exemplars. This is demonstrated in the ablation experiment (No. 3), where the removal of Dialogue Similarity results in a higher $W_d$ score, indicating a decrease in the alignment of segment boundaries. The similarity-guided in-context learning strategy is particularly beneficial in sparse and domain-specific settings, where the availability of relevant training data is limited. By leveraging semantically similar examples, the model can generalize better and achieve more accurate segmentation, thereby addressing the challenges of data sparsity and domain specificity.

3) The context-aware topic labeling module, which incorporates surrounding discourse, plays a critical role in generating more accurate and coherent topic annotations. Ours model achieves the best performance, with a $P_k$ score of 21.9 and a $W_d$ score of 33.9, indicating that all three components (Handshake, Dialogue Similarity, and Topic Generation) are necessary for optimal performance. The Topic Generation component ensures that the segments are semantically coherent, enhancing the overall quality of the topic segmentation. This is particularly important in public-channel conversations, where the context and discourse structure are complex and dynamic. The context-aware approach not only improves the accuracy of the topic labels but also makes the model more applicable to real-world scenarios, such as maritime radio and air traffic control communications.



\section{Conclusion}

This paper presents DASH-DTS, a structure-aware framework for Dialogue Topic Segmentation in public-channel conversations. It integrates handshake recognition, similarity-guided in-context learning, and context-aware topic labeling to effectively capture both structural and semantic cues in fragmented, high-stakes dialogues. In addition, to facilitate human-in-the-loop oversight in safety-critical settings, DASH-DTS outputs each predicted topic segment with an accompanying explanation and a confidence score, providing interpretability and trustworthiness estimation to support downstream decision-making.

Experiments on DialSeg711, Doc2Dial, and our newly collected public VHF channel dialog dataset \textbf{VHF-Dial} demonstrate that DASH-DTS achieves consistent improvements over state-of-the-art baselines, particularly in sparse-data and operational settings such as maritime VHF communication. Ablation studies validate the contribution of each module to the overall segmentation quality.

To support future research and real-world deployment, we release \textbf{VHF-Dial}, the first publicly available DTS dataset tailored to public-channel communication. 
Looking forward, we envision expanding this framework to other domains, such as air traffic control and emergency dispatch systems, where accurate, automated understanding of dialogue structure can directly improve situational awareness, reduce manual workload, and enable intelligent decision support in mission-critical environments.

\section{Limitation}
While DASH-DTS demonstrates significant improvements in topic segmentation for public-channel dialogues, several limitations remain. The framework's performance can be constrained by the sparsity and domain-specific nature of the data, particularly in specialized domains like maritime VHF communications. Additionally, the handshake recognition mechanism, though effective, may struggle with more subtle or complex interactional cues. The computational demands of the in-context learning component also pose a challenge for real-time and resource-constrained environments. Finally, enhancing the interpretability and explainability of the model's decisions is crucial for building trust and facilitating better integration with human operators in high-stakes communication settings. Future work should focus on addressing these limitations to further improve the robustness, efficiency, and practical applicability of DASH-DTS.

\section{Acknowledgments}
This research was funded by the Maritime Artificial Intelligence (AI) Research Project, supported under funding grant number SMI-2025-MTP-02 by the Singapore Maritime Institute (SMI).


\section*{Appendix}
\subsection{DTS Problem Formulation and Public-Channel Dataset}
\paragraph{Problem Statement}
Formally, Dialogue Topic Segmentation (DTS) can be framed as a structured prediction task that aims to detect latent topic boundaries within a dialogue \( D = \{u_1, u_2, ..., u_n\} \), where each utterance \( u_i \) is assigned to a contiguous topic segment \( T_j \). The output is a segmentation \( S = \{s_1, s_2, ..., s_k\} \), in which each \( s_j \subseteq D \) denotes a consecutive span of utterances with high topical coherence and minimal semantic drift. Unlike conventional utterance-level classification, DTS inherently requires modeling discourse-level dependencies, temporal progression, and transition phenomena that may span multiple turns. This is particularly challenging in task-oriented, public-channel environments, where lexical cohesion is weak, explicit discourse markers are absent, and topic changes often arise from pragmatic coordination rather than lexical shifts. Consequently, effective DTS systems must go beyond local similarity to incorporate structural, speaker-centric, and context-aware cues indicative of topical boundaries.

\paragraph{Dataset Overview and Application scenarios }
A number of established datasets \cite{xie2021tiage} have advanced research in Dialogue Topic Segmentation, including \textbf{DialSeg711} \cite{xu2020topic} and \textbf{Doc2Dial} \cite{feng2020doc2dial}. These datasets are primarily drawn from structured interviews, multi-party meetings, or online discussions, where utterances tend to be longer, well-formed, and contextually rich. While effective for modeling discourse in open-domain or cooperative settings, such datasets fall short in representing the terse, fragmented, and pragmatically complex nature of public-channel communications.

We are motivated by pressing real-world needs observed in industrial maritime regulatory practice. In particular, a new dataset was purposefully collected and constructed to support DTS under public-channel communication scenarios—specifically, maritime VHF radio dialogues, namely \textbf{VHF-Dial}. In real-world coastal monitoring zones, massive volumes of open-channel voice communications take place between ships and maritime regulators (and, to a lesser extent, between ships themselves), following international conventions on shared frequency use. These conversations are mission-critical, highly dynamic, and span a wide array of tasks ranging from port entry coordination to emergent risk mitigation.

However, structuring such data at scale is almost impossible through manual means. Of even greater concern, near-miss events—non-accidental but high-risk interactions—are frequently omitted from formal reporting pipelines due to the absence of physical consequences. These events are valuable for proactive safety assessment, yet their analysis demands hours of manual transcription and annotation. This imposes high labor costs and limits the ability of current digital oversight systems to learn from such rich but latent sources.

We envision DTS as a fundamental building block to alleviate this bottleneck. If we can accurately and automatically identify topic transitions within public-channel VHF conversations, it would unlock structured behavioral insights at scale and provide regulatory authorities with improved situational awareness and data-driven decision support.

\end{document}